\documentclass[journal]{IEEEtran}
\usepackage{amsmath,amssymb,amsfonts}
\usepackage{algorithmic}
\usepackage{graphicx}
\usepackage{textcomp}
\usepackage{xcolor,soul}
\usepackage{url}

\usepackage{multirow}
\usepackage{comment}
\usepackage[linesnumbered,ruled,vlined]{algorithm2e}
\usepackage[export]{adjustbox}
\usepackage{listings}
\usepackage{enumitem}
\usepackage{subcaption}
\usepackage[hidelinks]{hyperref}
\begin{document}

\title{Provenance of AI-Generated Images: A Vector Similarity and Blockchain-based Approach}
% \shorttitle{Embedding-Based Detection of AI-Generated Images}
% \researchtype{Completed Research Papers}
% \track{Human-computer Interaction}

\author{
    \IEEEauthorblockN{
        Jitendra Sharma\IEEEauthorrefmark{1}, 
        Arthur Carvalho\IEEEauthorrefmark{2},
        Suman Bhunia\IEEEauthorrefmark{1}
    } \\
    \IEEEauthorblockA{\IEEEauthorrefmark{1}
    Department of Computer Science and Software Engineering, Miami University, Oxford, Ohio}\\
    \IEEEauthorblockA{\IEEEauthorrefmark{2}
    Farmer School of Business, Miami University, Oxford, Ohio
    }\\
    \IEEEauthorblockA{
    Email:  \url{sharmaj2@miamioh.edu}, arthur.carvalho@miamioh.edu, bhunias@miamioh.edu
    }
}

\maketitle
\begin{abstract}
Rapid advancement in generative AI and large language models (LLMs) has enabled the generation of highly realistic and contextually relevant digital content. LLMs such as ChatGPT with DALL-E integration and Stable Diffusion techniques can produce images that are often indistinguishable from those created by humans, which poses challenges for digital content authentication. Verifying the integrity and origin of digital data to ensure it remains unaltered and genuine is crucial to maintaining trust and legality in digital media. In this paper, we propose an embedding-based AI image detection framework that utilizes image embeddings and a vector similarity to distinguish AI-generated images from real (human-created) ones. Our methodology is built on the hypothesis that AI-generated images demonstrate closer embedding proximity to other AI-generated content, while human-created images cluster similarly within their domain. To validate this hypothesis, we developed a system that processes a diverse dataset of AI and human-generated images through five benchmark embedding models.  Extensive experimentation demonstrates the robustness of our approach, and our results confirm that moderate to high perturbations minimally impact the embedding signatures, with perturbed images maintaining close similarity matches to their original versions. Our solution provides a generalizable framework for AI-generated image detection that balances accuracy with computational efficiency.

\end{abstract}

\section{Introduction}\label{sec:intro}
Generative AI and large language models (LLMs), such as ChatGPT~\cite{openai_chatgpt}, have rapidly advanced, enabling the creation of human-like text and highly realistic images from simple prompts. Diffusion models like DALL-E, Imagen, Stable Diffusion, and Midjourney have transformed image generation, producing diverse outputs in seconds~\cite{bengesi2024advancements,ramesh2021zero}. As these models grow more sophisticated, their outputs become increasingly difficult to distinguish from human-made content, creating new challenges for detecting AI-generated material.

The proliferation of AI-generated content poses serious concerns across creative and informational domains, as these models can mimic human artistic styles~\cite{winn2023frontiers,iyer2023generative} and potentially undermine the value of original works~\cite{madhu2023survey}. High-profile incidents have already sparked debate about AI’s role in art~\cite{Roose2022} and its potential to spread disinformation through convincingly fake imagery~\cite{pbs2023fakeimages}. Although progress has been made in AI-generated image detection, most existing methods are tailored to specific models (e.g., GANs or diffusion models) and often fail to generalize across newer versions~\cite{aiornot_petapixel, sandotra2024}. Recent studies also highlight the limitations of watermarking techniques, which remain vulnerable to evasion~\cite{jiang2023evading}, underscoring the need for more adaptable and robust detection approaches.

Several recent studies have attempted to improve detection methods by exploring forensic traces, embedding-based classification, and hybrid strategies. Corvi et al.~\cite{corvi2023detection} analyzed synthetic image artifacts left by diffusion models and observed that detector performance varied widely across different generators, especially after common post-processing operations like compression and resizing. Ojha et al.~\cite{ojha2023towards} proposed using the CLIP-ViT embedding space for nearest-neighbor and linear classification, demonstrating improved accuracy on unseen diffusion models, although their method was less robust to heavy manipulations. Baraheem et al.~\cite{baraheem2023ai} achieved 100\% accuracy on a custom dataset using EfficientNetB4 and Class Activation Maps, but the method’s dependency on curated training data limits its generalizability. Martin et al.~\cite{martin2023detection} combined forensic methods like PRNU and Error Level Analysis with CNNs, achieving strong precision but with limitations in image format support and adversarial robustness. These studies highlight the need for more generalizable and manipulation-resilient detection systems.

Motivated by these limitations, our work introduces a detection system that avoids heavy model training and instead leverages robust image embeddings and vector similarity to identify synthetic content. To address these challenges, we present the EmbedAIDetect system—a lightweight, training-free detection mechanism that determines whether an image is AI-generated using embedding similarity. Our approach leverages a Vision Transformer-based embedding model in combination with a vector database to compare incoming images against a reference set. To enhance transparency and tamper-resistance, the system also integrates blockchain-based verification. The solution is deployed as a web application with a user-friendly interface. The main contributions of this research are:

\begin{itemize}
    \item Provides a resource-efficient, training-less AI image detection system that utilizes a similarity score to determine the authenticity of images.
    \item Identifies and implements a suitable embedding technique that is generalizable and resistant to intentional manipulations and post-adversarial tampering attacks.
    \item Integrate vector databases with embedding model and blockchain to maintain a tamper proof record of embeddings enhancing integrity of the detection system for real-time purposes.
\end{itemize}
\section{System Design}\label{sec:system}
\begin{figure}[ht]
\centering
\includegraphics[width=\columnwidth]{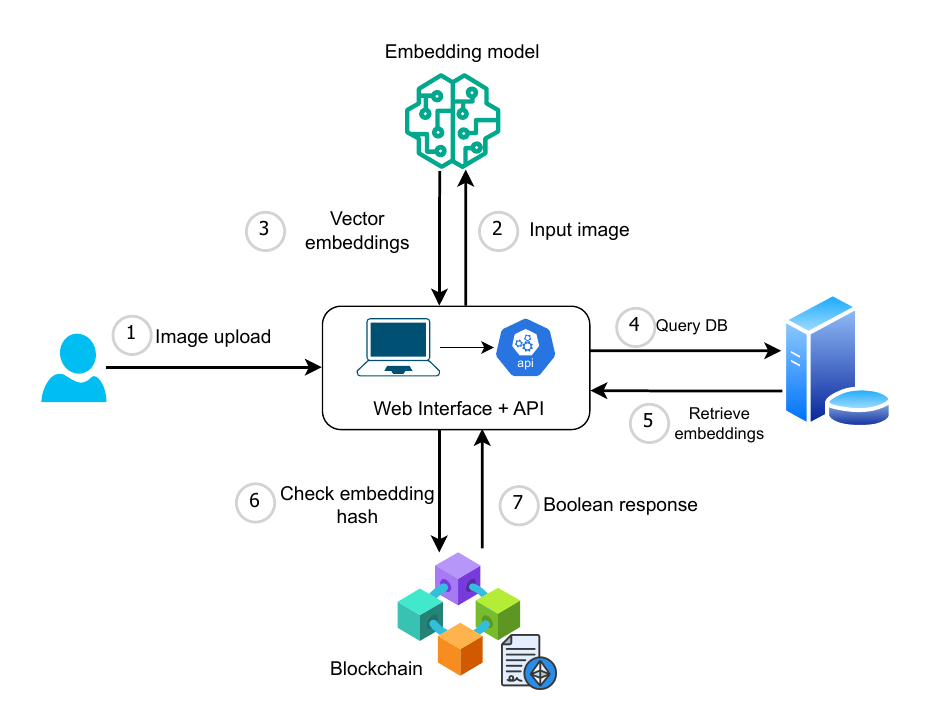}
\caption{System architecture for EmbedAIDetect}
\label{fig:system-architecture}
\end{figure}

The paper presents the framework designed to integrate vector databases with the embedding model and a user-friendly interface. The framework implements a step-by-step pipeline for analyzing whether an unseen (uploaded) image might be AI-generated. Figure \ref{fig:system-architecture} illustrates the architecture design for the system. It provides a user interface to upload a photo and check the results. The results are based on the embedding vectors that are stored in the vector database. A detailed walk-through of each component and its interaction is described in the following.

\subsection{System Components}

\subsubsection{Embedding Model}
The system uses a Vision Transformer (ViT) model from the DINOv2 framework~\cite{oquab2023dinov2}, implemented using Hugging Face’s \texttt{transformers} library. A custom embedding function is built using \texttt{AutoModel} and \texttt{AutoProcessor}, which preprocesses input images into tensors for the model to interpret. The model runs on a GPU and outputs patch-level hidden states, which are averaged to produce a single high-dimensional vector embedding. This embedding is moved to CPU memory, converted into a NumPy array, and flattened into a Python list to ensure compatibility with the vector databases for similarity search and storage.

\subsubsection{Vector Database}
We use \textit{ChromaDB}~\cite{chroma2025docs}, an open-source vector database optimized for similarity search in high-dimensional spaces. The system initializes it using \texttt{PersistentClient}, with collections configured for AI-generated and human-generated image embeddings. Each collection is linked to a custom embedding function based on the DINOv2 model and uses cosine distance as the similarity metric. When a user uploads an image, its embedding is computed and used to query the database for the top-$k$ closest matches. ChromaDB handles both embedding storage and retrieval, while a backend script processes the results and displays them to the user via the Streamlit interface.

\subsubsection{Blockchain Setup}
Blockchain provides a decentralized and tamper-resistant way to store data, particularly useful for verifiable systems~\cite{shrestha2024case}. In this project, the Ethereum network was selected for its mature ecosystem and support for smart contracts. Two Solidity-based contracts—\texttt{HashStorageAI} and \texttt{HashStorageHuman}—were developed with identical logic, each exposing functions to store and verify 256-bit hashes of image embeddings. These contracts were tested using Hardhat and deployed to the Sepolia testnet, selected for its Proof-of-Stake consensus and compatibility with Ethereum’s mainnet. Backend integration is managed through a custom Python wrapper using the Web3.py library, which enables interaction with the contracts and provides application-level methods like \texttt{store\_hash()} and \texttt{check\_hash\_exists()}. MetaMask handles secure transaction signing and private key management during blockchain operations.

\subsubsection{Web Application}
The web application serves as the user-facing layer that integrates the embedding model, vector database, and blockchain verification components into a unified workflow. Built using \textit{Streamlit}~\cite{streamlit2025docs}, it provides a simple and responsive interface through which users can upload images for analysis. Once an image is uploaded, the system processes it using the DINOv2 model to generate an embedding, which is then compared against stored vectors in ChromaDB using cosine similarity to estimate whether the image is AI-generated or human-generated. Simultaneously, a 256-bit hash of the embedding is computed and checked against Ethereum smart contracts to determine if it has been previously recorded on-chain. The application consolidates the similarity-based classification and blockchain verification into a single output, presenting users with both the prediction result and its verifiability in a clear and accessible manner.

\subsection{System Implementation}\label{sec:implementation}

The EmbedAIDetect system is developed through three distinct frameworks, each building upon the capabilities of the previous one. This progressive development approach demonstrates how each framework was conceptualized, implemented, and refined to advance toward the final integrated solution. Each version introduces new components—starting from a blockchain-only setup, to incorporating a vector database for similarity search, and finally combining both for verification and classification. Detailed flowcharts and explanations for each framework are provided in the following sections to illustrate the application logic and architectural decisions. 

\subsubsection{Framework 1: Blockchain only}
\begin{figure}
    \centering
    \begin{subfigure}[t]{0.45\textwidth}
        \includegraphics[width=\linewidth]{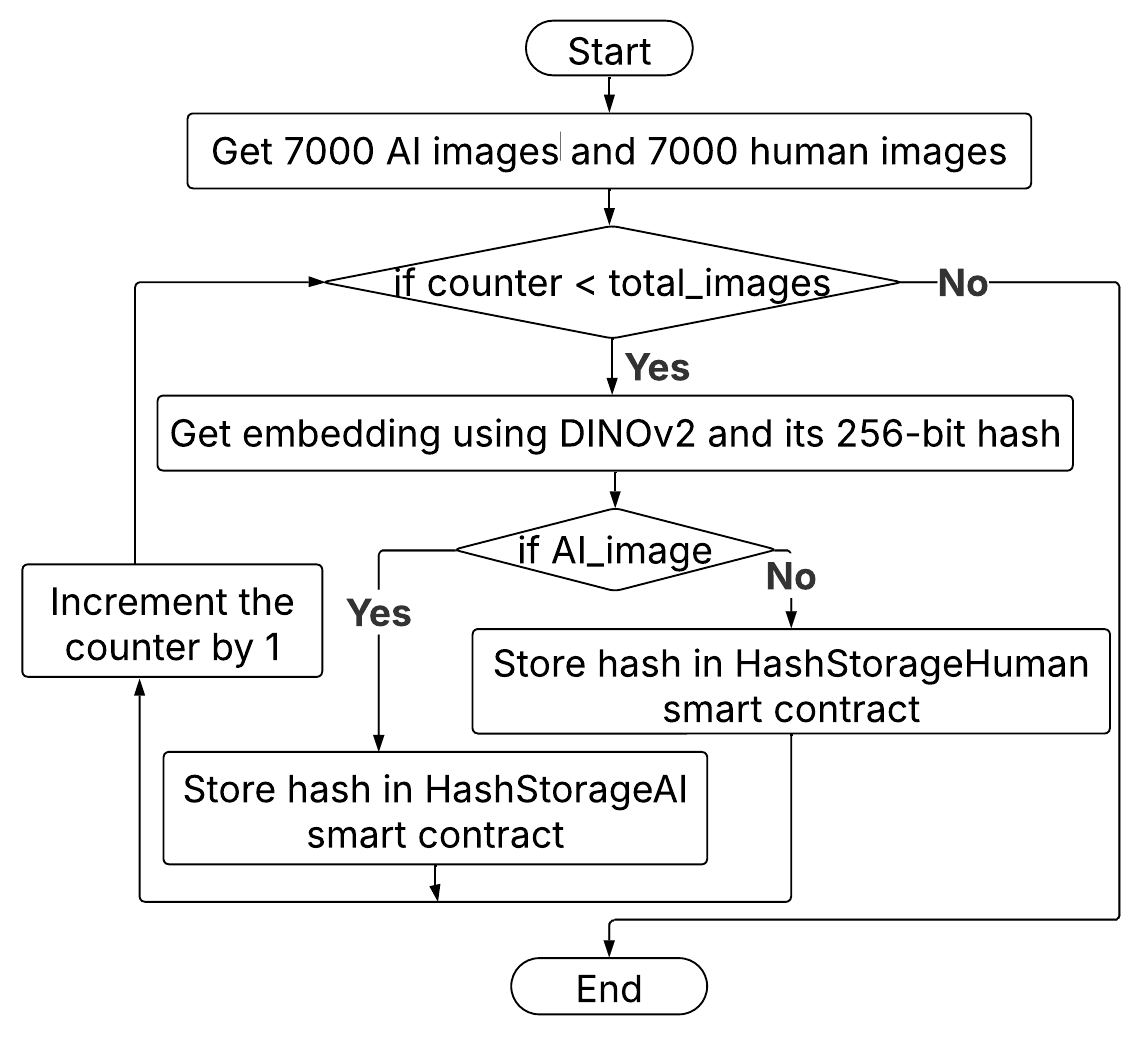}
        \caption{Hash storage in smart contracts setup}
    \end{subfigure}
    \begin{subfigure}[t]{0.45\textwidth}
        \includegraphics[width=\linewidth]{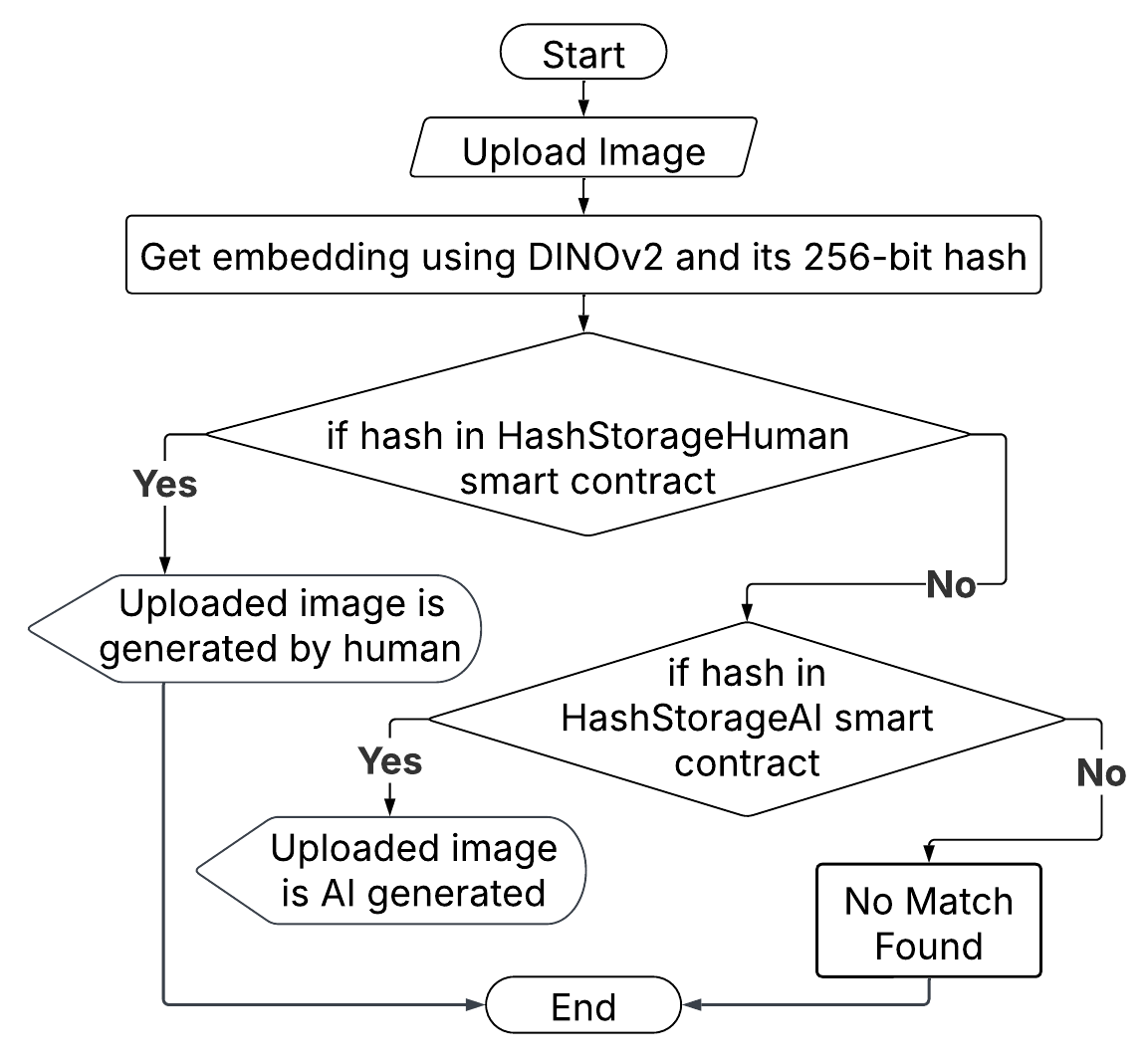}
        \caption{Application flowchart illustrating the image upload process and hash-based classification using smart contracts}
    \end{subfigure}
    \caption{Application setup and data flowchart for blockchain only framework of EmbedAIDetect}
    \label{fig:framework1}
\end{figure}

The first approach as shown in Figure~\ref{fig:framework1}, the system leverages blockchain technology to store and verify image embeddings using smart contracts. Two separate smart contracts are deployed: one named HashStorageAI for AI-generated image hashes, and another named HashStorageHuman for human-generated image hashes. The image embeddings are generated using the DINOv2 model, and these embeddings are then converted into 256-bit hash values. These hashes are immutably stored on the blockchain to serve as reference data for later verification.

The application begins by initializing arrays for storing AI and human image data. A dataset of 7,000 AI-generated images and 7,000 human-generated images is processed. For each image, an embedding is generated using the DINOv2 model, which is then hashed into a 256-bit value. The image known to be AI-generated, their hash is stored in the HashStorageAI smart contract; otherwise, it’s stored in the HashStorageHuman contract. When a new image is uploaded for verification, the system generates its embedding and hash, then checks for a match in both smart contracts. A match in HashStorageAI indicates the image is AI-generated, while a match in HashStorageHuman indicates it’s human-generated. If no match is found, the result is not determined.

\subsubsection{Framework 2: Vector Database only}

\begin{figure}
    \centering
    \begin{subfigure}[t]{0.45\textwidth}
        \includegraphics[width=\linewidth]{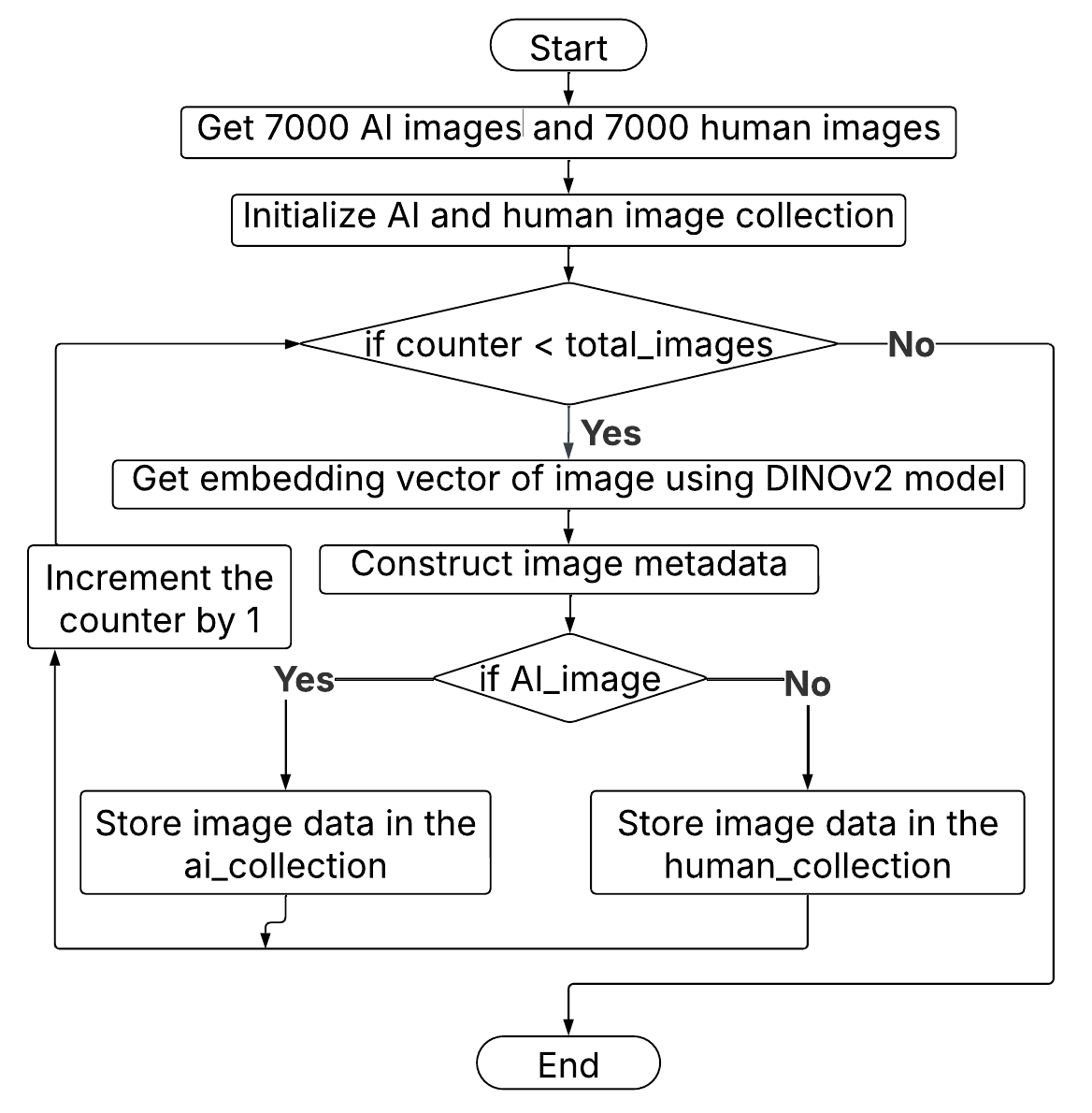}
        \caption{Embeddings storage in separate collections in ChromaDB vector database flowchart}
    \end{subfigure}
    \begin{subfigure}[t]{0.45\textwidth}
        \includegraphics[width=\linewidth]{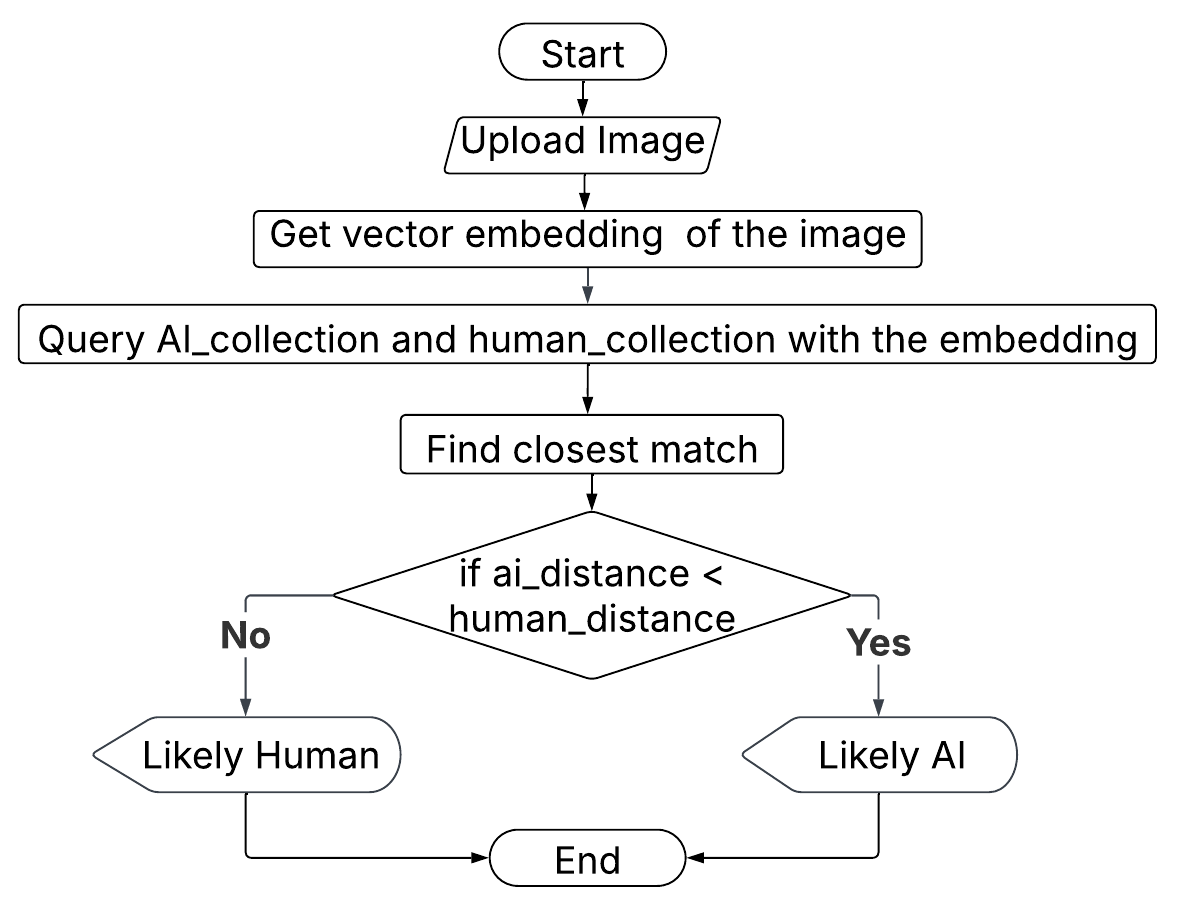}
        \caption{Image classification process using vector 
similarity search in ChromaDB}
    \end{subfigure}
    \caption{Application setup and data flowchart for the vector database only framework of EmbedAIDetect }
    \label{fig:framework2}
\end{figure}

The second approach as shown in Figure~\ref{fig:framework2} utilizes a vector database, specifically ChromaDB, to store and compare image embeddings. Two separate collections are created: one for AI-generated images and another for human-generated images. The system uses the DINOv2 model to extract vector embeddings from images. Each embedding, along with metadata, is stored in the corresponding collection based on the image's classification as AI or human. This approach facilitates efficient similarity searches based on vector distances during inference.

The application starts by initializing two collections in ChromaDB: one for AI images and one for human images. A dataset of 7,000 AI-generated and 7,000 human-generated images is processed. Each image is converted into a vector embedding using the DINOv2 model, and this embedding, along with relevant metadata, is added to the appropriate collection based on the image type. When a new image is uploaded, the system generates its vector embedding and performs similarity queries against both collections. It then compares the distances: if the embedding is closer to the AI collection, the image is classified as likely AI-generated; otherwise, it's labeled as likely human-generated.

\subsubsection{Framework 3: Hybrid Approach}

\begin{figure}
    \centering
    \begin{subfigure}[t]{0.45\textwidth}
        \includegraphics[width=\linewidth]{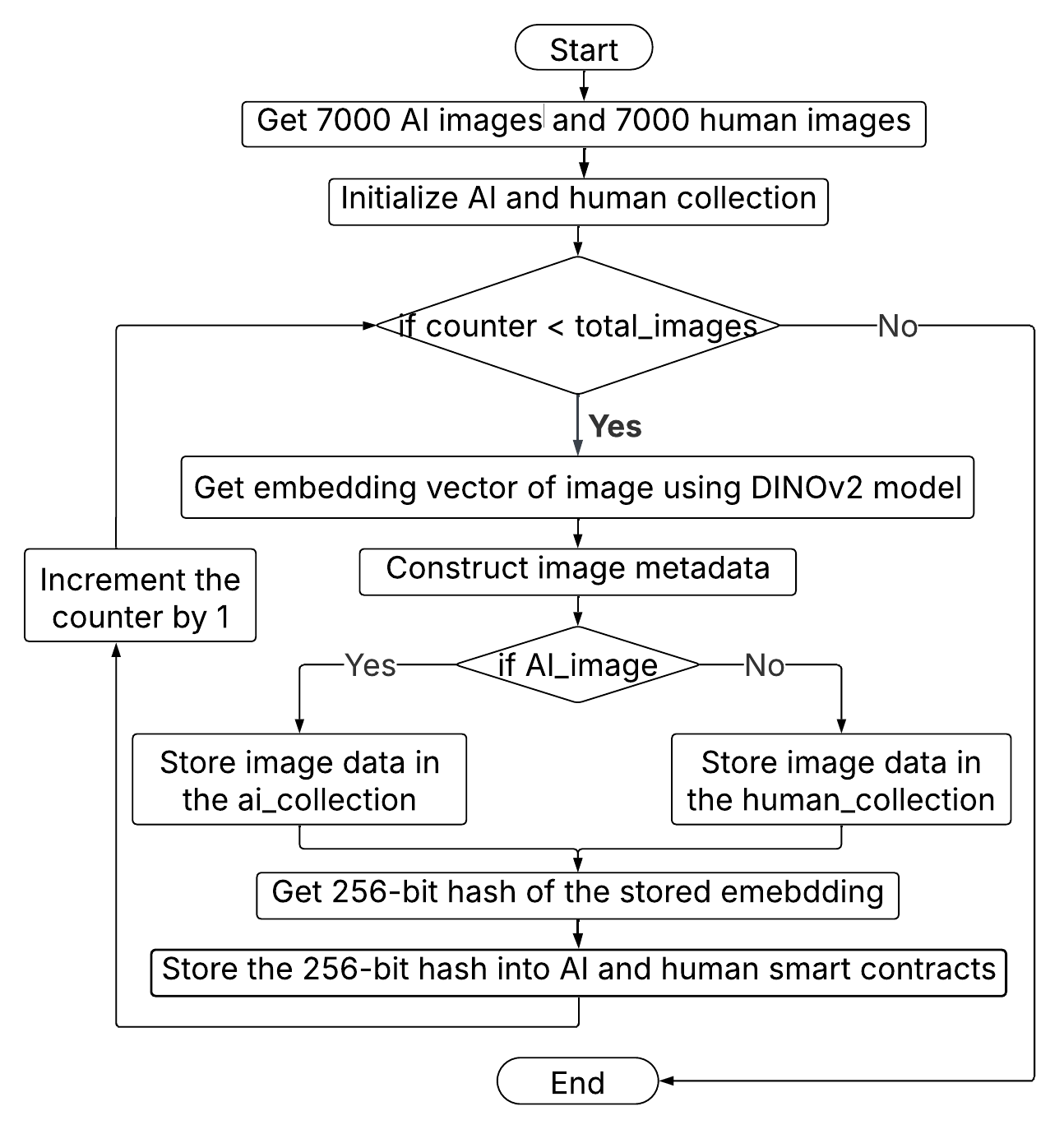}
        \caption{Embeddings and 256-bit hash storage setup flowchart}
    \end{subfigure}
    \begin{subfigure}[t]{0.45\textwidth}
        \includegraphics[width=\linewidth]{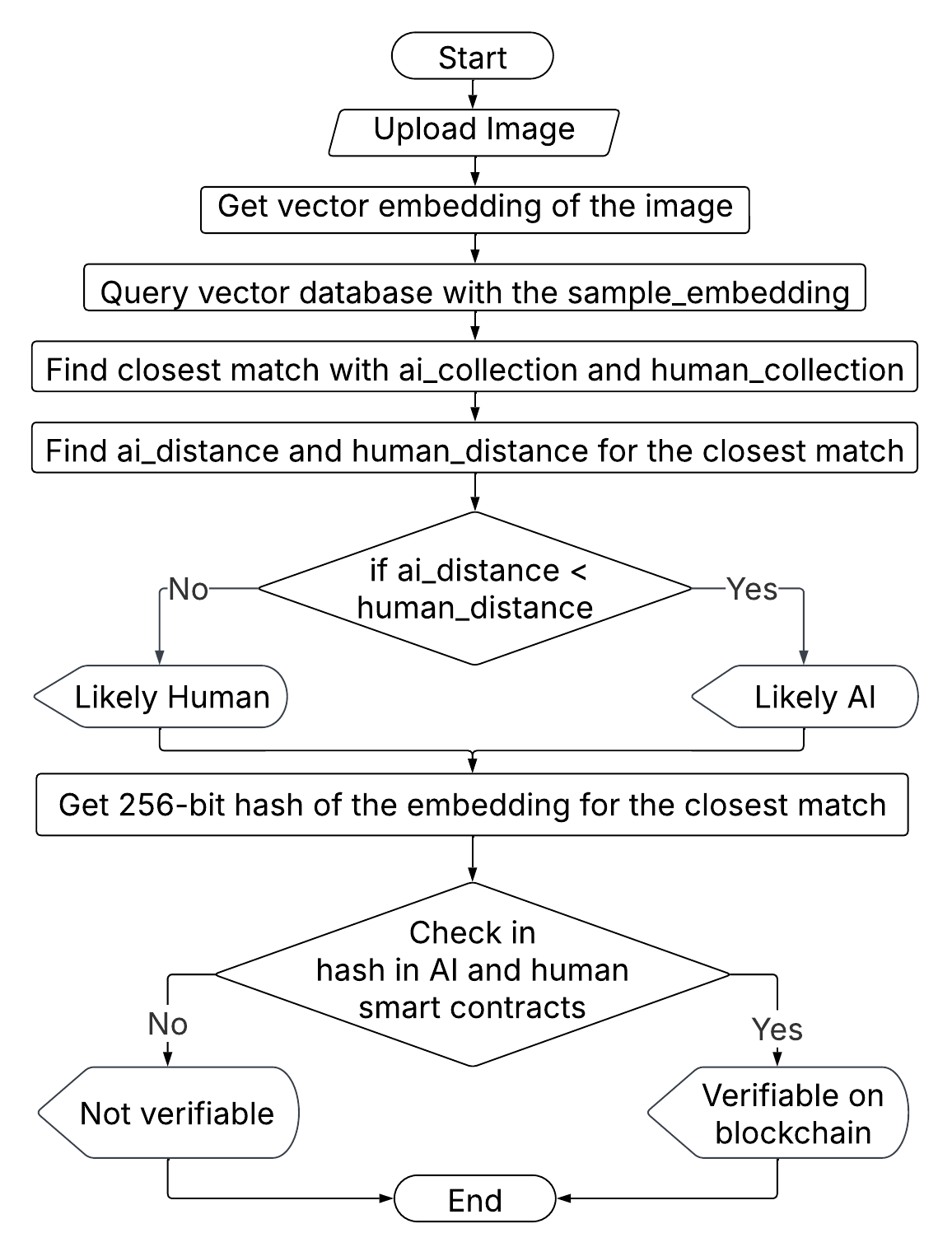}
        \caption{Combining vector similarity search and blockchain-based verification to classify and authenticate uploaded images }
    \end{subfigure}
    \caption{Application setup and data flowchart for the hybrid framework of EmbedAIDetect}
    \label{fig:framework3}
\end{figure}

The third approach, which is the hybrid method shown in Figure~\ref{fig:framework3} combines blockchain-based verification with a vector database similarity search to enhance both the accuracy and integrity of AI-generated image detection. The setup begins by initializing two collections in ChromaDB: one for AI-generated images and the other for human-generated images. Simultaneously, two smart contracts—HashStorageAI and HashStorageHuman—are deployed on the blockchain to store 256-bit hashes of image embeddings. A dataset of 7,000 AI-generated and 7,000 human-generated images is processed: each image is passed through the DINOv2 model to extract an embedding. This embedding is stored as a vector in the respective ChromaDB collection along with metadata, and also hashed and stored immutably on the blockchain in the corresponding smart contract.

When a new image is uploaded, its embedding is generated using the DINOv2 model. This embedding is used to query both ChromaDB collections to retrieve the closest matching vectors—one from the AI set and one from the human set. The distances between the uploaded image and each match are compared: if the image is closer to the AI collection, it is classified as likely AI-generated; otherwise, it is labeled as likely human-generated. In parallel, the 256-bit hash of the image embedding is computed and checked against both HashStorageAI and HashStorageHuman. If a match is found in either contract, the classification is deemed verifiable on-chain. If no match exists, the result is not verifiable, though the classification can still be inferred through vector similarity.

\section{Evaluation}\label{sec:eval}
To rigorously assess the performance and robustness of EmbedAIDetect, we designed a comprehensive experimental framework focused on evaluating the role of vector embeddings in detecting AI-generated images. This framework investigates the identification and implementation of an embedding technique that is generalizable and resilient to deliberate manipulations. The experimental setup consists of three key studies: first vector similarity-based classification for detection, second that benchmarks various embedding models to determine their effectiveness under adversarial conditions, and lastly, estimation of gas usage while using smart contracts.

All experiments were carried out in a high performance computing environment equipped with an Intel Core i9-14900K CPU, 66 GB of RAM, and an NVIDIA GeForce RTX 4090 GPU to ensure efficient handling of computationally intensive tasks. To evaluate the blockchain integration aspects of the system, we used Hardhat, a widely adopted Ethereum development framework, to simulate gas usage and smart contract interactions in a controlled environment.

\subsection{Vector similarity based classification}

The goal of this experiment is to evaluate the feasibility and accuracy of using vector embeddings and cosine similarity metrics to differentiate between AI-generated and human-captured images. Whether AI-generated or human-created, both images exhibit unique patterns in their embeddings when passed through a pre-trained image classification model.

We constructed a diverse data set that encompasses real and AI-generated images, except human faces. It consists of 9,000 AI-generated images and 6,074 human-created art images. The AI images are created using the Stable Diffusion 3.5 Medium model provided by Stability AI and is available on the Hugging Face platform. The images are generated at a fixed resolution of $512 \times 512$ to balance quality and generation time. The AI images are generated using systematically crafted prompts combining various themes, subjects, and styles, while the human dataset is sourced from the Kaggle art-images collection, carefully curated to remove corrupted and duplicate entries.

Below are representative examples of prompts used to generate AI images:
\begin{itemize}
    \item ``A tropical floating islands in glitch art style"
    \item ``A biotechnological deep space in photorealistic style''
    \item ``A time travel rainforest canopy in psychedelic art style''
    \item ``A neon concert hall in low poly style''
    \item ``A mystical art studio in neon style''
\end{itemize}

The sample images are shown in Figure \ref{fig:sample-dataset} below, illustrate the dataset's nature and types of manipulations applied, providing insight into how the system evaluates and identifies AI-generated content.

\begin{figure*}
    \centering
    
    % First row of images
    \begin{subfigure}[t]{0.10\textwidth}
        \includegraphics[width=\linewidth]{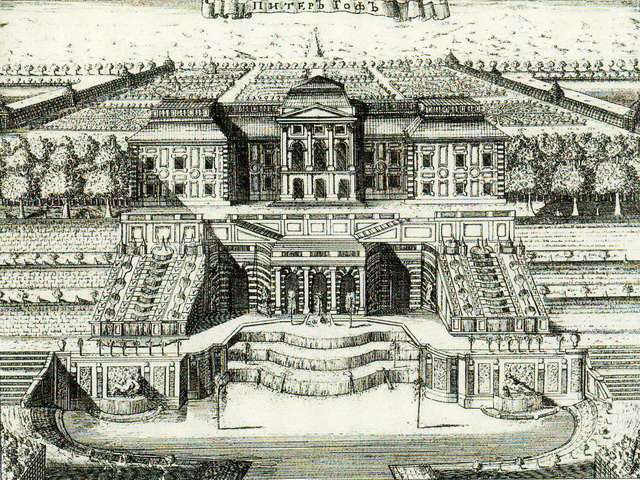}
    \end{subfigure}
    \begin{subfigure}[t]{0.10\textwidth}
        \includegraphics[width=\linewidth]{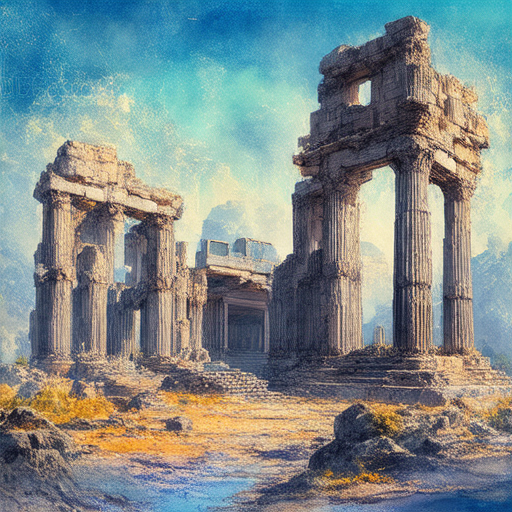}
    \end{subfigure}
    \begin{subfigure}[t]{0.10\textwidth}
        \includegraphics[width=\linewidth]{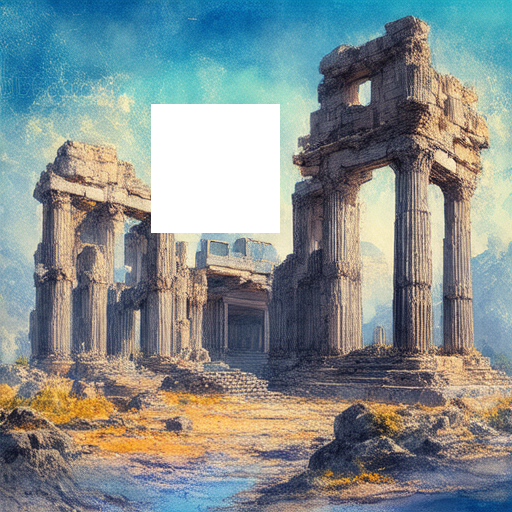}
    \end{subfigure}
    \begin{subfigure}[t]{0.10\textwidth}
        \includegraphics[width=\linewidth]{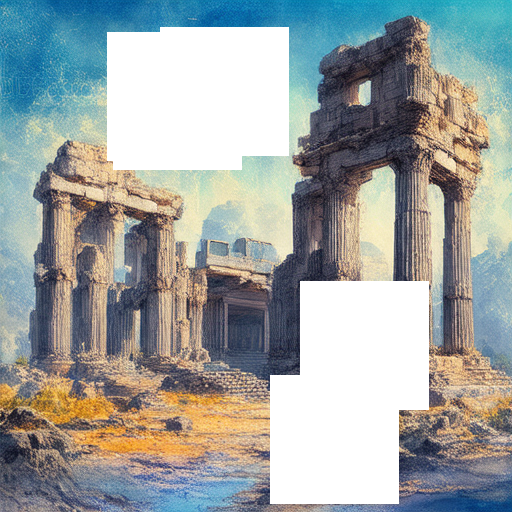}
    \end{subfigure}
    \begin{subfigure}[t]{0.10\textwidth}
        \includegraphics[width=\linewidth]{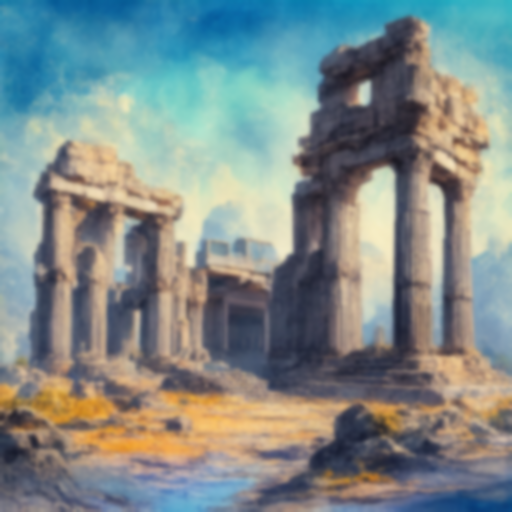}
    \end{subfigure}
    \begin{subfigure}[t]{0.10\textwidth}
        \includegraphics[width=\linewidth]{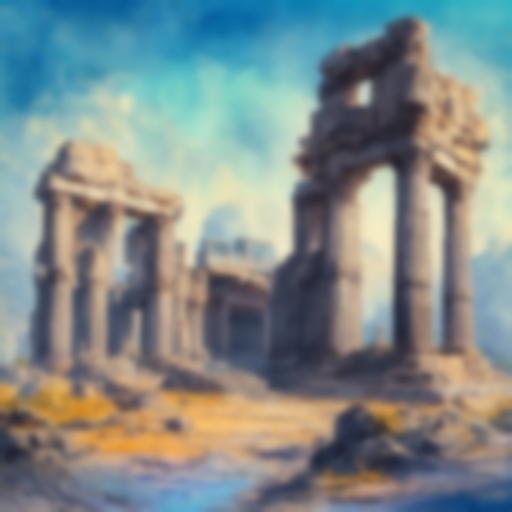}
    \end{subfigure}
    \begin{subfigure}[t]{0.10\textwidth}
        \includegraphics[width=\linewidth]{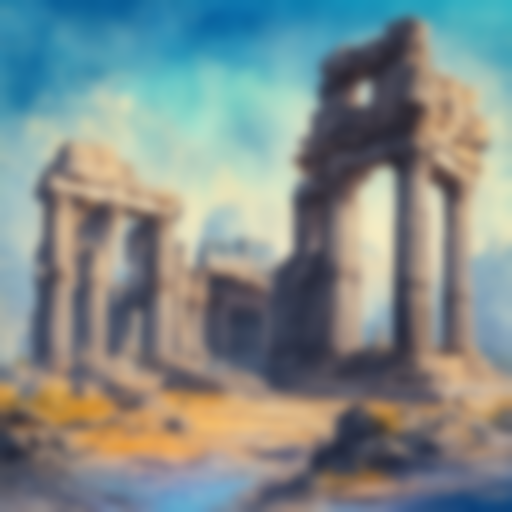}
    \end{subfigure}
    \begin{subfigure}[t]{0.10\textwidth}
        \includegraphics[width=\linewidth]{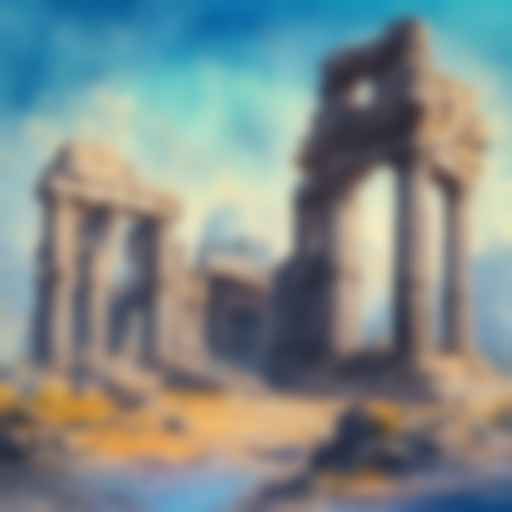}
    \end{subfigure}

    % Second row of images
    \begin{subfigure}[t]{0.10\textwidth}
        \includegraphics[width=\linewidth]{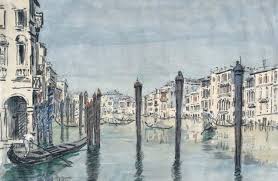}
    \end{subfigure}
    \begin{subfigure}[t]{0.10\textwidth}
        \includegraphics[width=\linewidth]{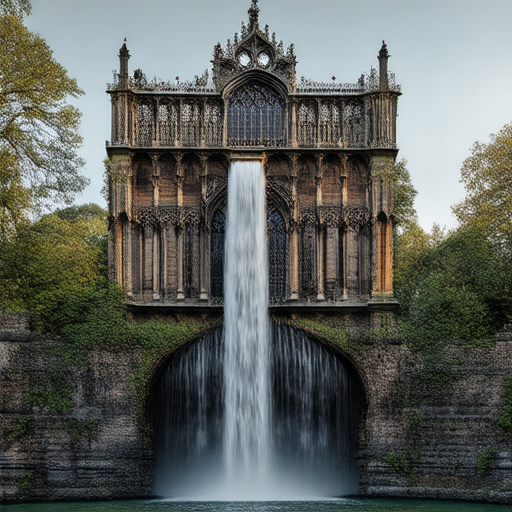}
    \end{subfigure}
    \begin{subfigure}[t]{0.10\textwidth}
        \includegraphics[width=\linewidth]{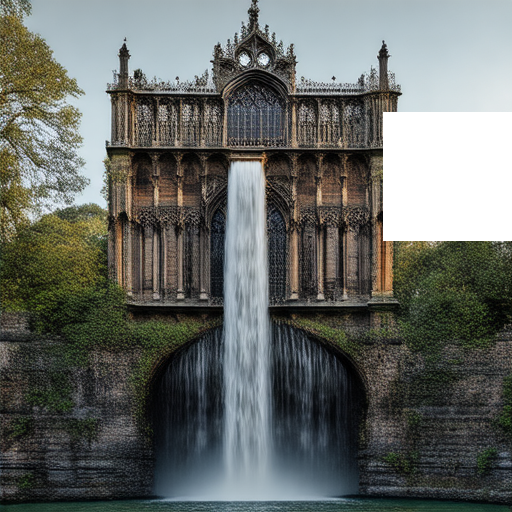}
    \end{subfigure}
    \begin{subfigure}[t]{0.10\textwidth}
        \includegraphics[width=\linewidth]{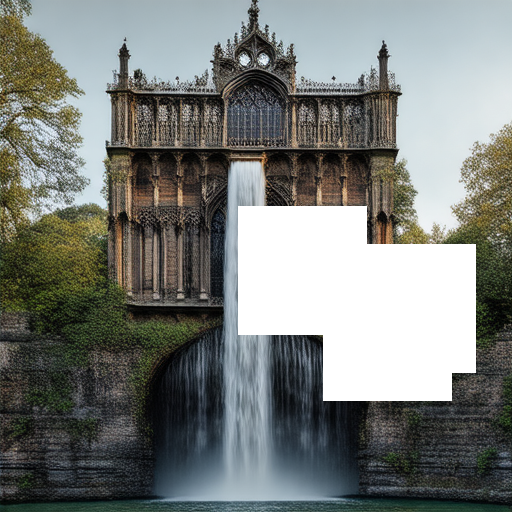}
    \end{subfigure}
    \begin{subfigure}[t]{0.10\textwidth}
        \includegraphics[width=\linewidth]{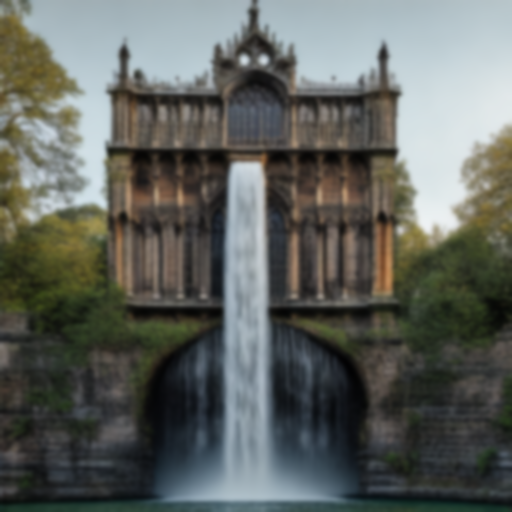}
    \end{subfigure}
    \begin{subfigure}[t]{0.10\textwidth}
        \includegraphics[width=\linewidth]{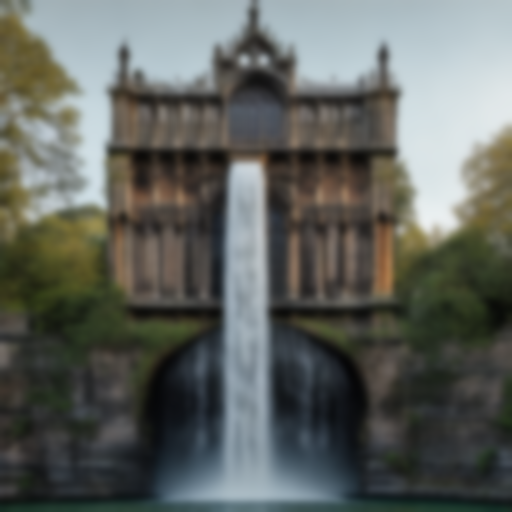}
    \end{subfigure}
    \begin{subfigure}[t]{0.10\textwidth}
        \includegraphics[width=\linewidth]{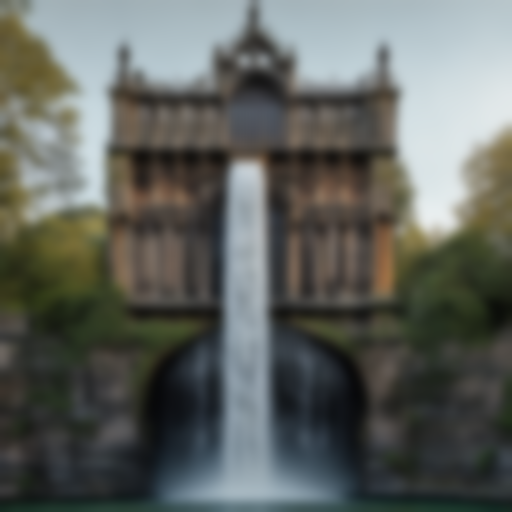}
    \end{subfigure}
    \begin{subfigure}[t]{0.10\textwidth}
        \includegraphics[width=\linewidth]{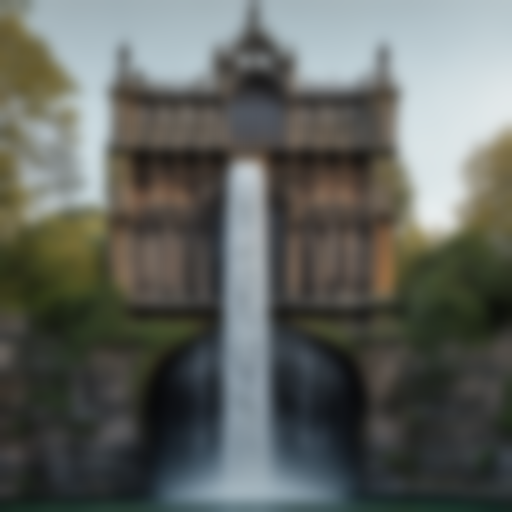}
    \end{subfigure}

    % Third row of images
    \begin{subfigure}[t]{0.10\textwidth}
        \includegraphics[width=\linewidth]{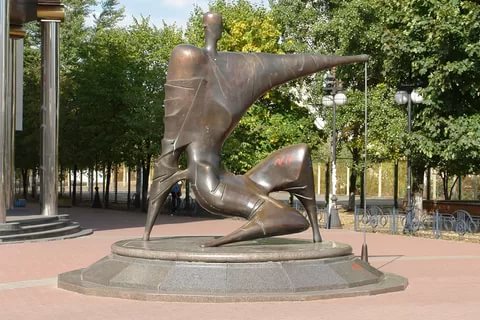}
        \caption{Human art image}
    \end{subfigure}
    \begin{subfigure}[t]{0.10\textwidth}
        \includegraphics[width=\linewidth]{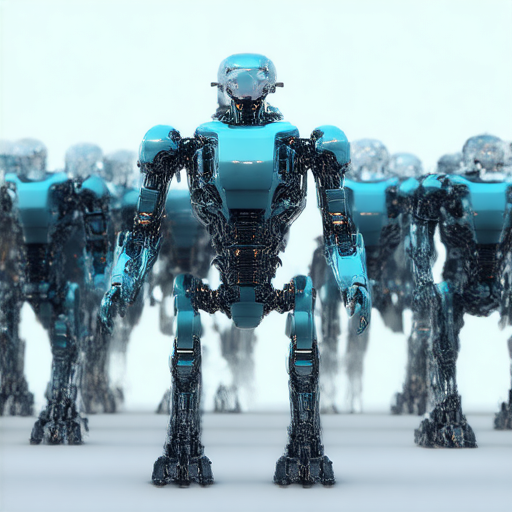}
        \caption{AI image}
    \end{subfigure}
    \begin{subfigure}[t]{0.10\textwidth}
        \includegraphics[width=\linewidth]{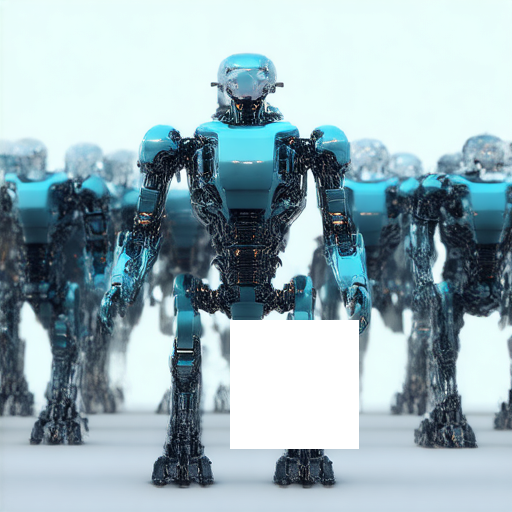}
        \caption{One patch}
    \end{subfigure}
    \begin{subfigure}[t]{0.10\textwidth}
        \includegraphics[width=\linewidth]{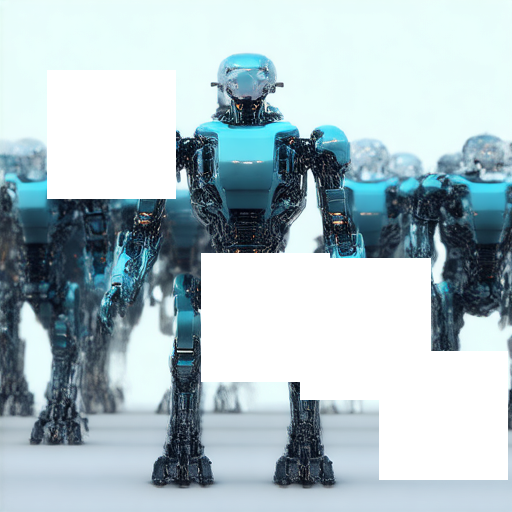}
        \caption{Multiple patch}
    \end{subfigure}
    \begin{subfigure}[t]{0.10\textwidth}
        \includegraphics[width=\linewidth]{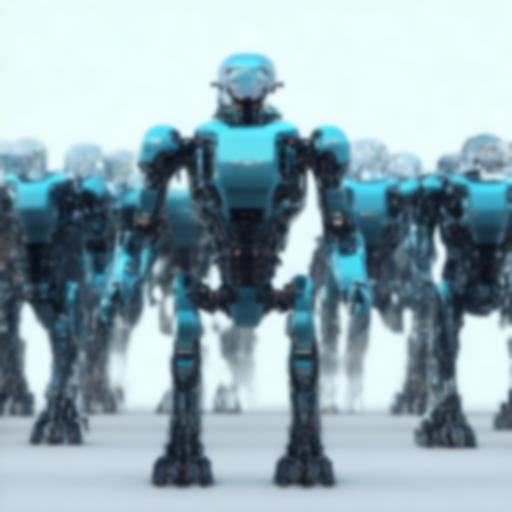}
        \caption{20\% blur}
    \end{subfigure}
    \begin{subfigure}[t]{0.10\textwidth}
        \includegraphics[width=\linewidth]{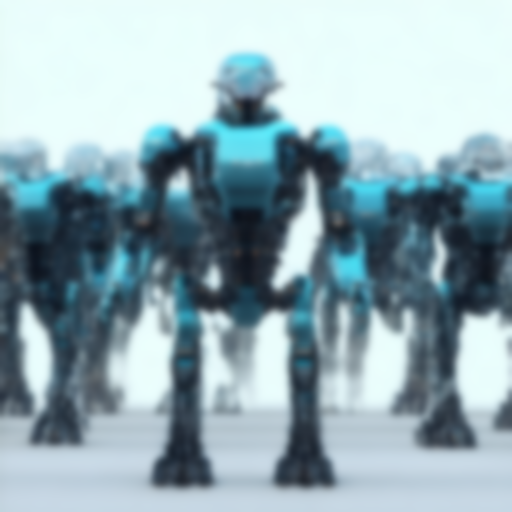}
        \caption{40\% blur}
    \end{subfigure}
    \begin{subfigure}[t]{0.10\textwidth}
        \includegraphics[width=\linewidth]{Figures/sample_dataset/blur60_2.png}
        \caption{60\% blur}
    \end{subfigure}
    \begin{subfigure}[t]{0.10\textwidth}
        \includegraphics[width=\linewidth]{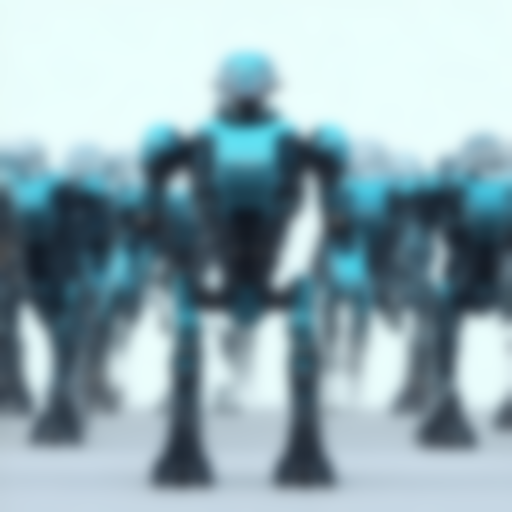}
        \caption{80\% blur}
    \end{subfigure}

    \caption{Sample images from human-art dataset and AI-generated image dataset with various modifications}
    \label{fig:sample-dataset}
\end{figure*}

Our methodology implements a novel classification pipeline utilizing vector embeddings extracted through an embedding model. We stored vector embeddings of generated AI images and human-created art images in the Pinecone vector database. Separated partitions are created for training and testing datasets split in a 4:1 ratio. The closest distance for each testing image in both the AI image training set and the human training set is calculated and stored in spreadsheets. The nearest distance is calculated using the similarity search and cosine distance as a metric. The experiment is done repeatedly using five pre-trained models: OpenAI CLIP (512-dimensional embeddings), Google ViT (768-dimensional embeddings), Facebook DINOv2 (768-dimensional embeddings), Microsoft ResNet-50 (2048-dimensional embeddings), and Apple AIMv2 (1024-dimensional embeddings). We used separate indexes for each embedding model with name-space partitioning for training and testing data sets to store the vector embeddings.
The two spreadsheets one of ai image test set and other of human image test set are used to calculate confusion matrix based on the classification algorithm below, which is the main hypothesis the detection system is based on.

\begin{equation}
\text{AI\_or\_not}(x) = 
\begin{cases} 
1 & \text{if } d(x, \mathcal{A}) \leq d(x, \mathcal{H}) \\
0 & \text{if } d(x, \mathcal{A}) > d(x, \mathcal{H})
\end{cases}
\end{equation}

\text{where;}
\begin{align*}
x &: \text{supplied image embedding vector} \\
\mathcal{A} &: \text{set of AI-generated image embeddings} \\
\mathcal{H} &: \text{set of human-generated image embeddings} \\
d(x, S) &= \min_{s \in S} \{\text{cosine\_distance}(x, s)\} \\
\end{align*}

The experimental results demonstrated exceptional performance across all models, as shown in Table \ref{tab:performance}. CLIP achieved the highest overall accuracy at 99.51\% with near-perfect precision (0.9935) and recall (0.998). AIMv2 demonstrated consistent performance across all metrics, while ResNet-50 showed lower recall despite maintaining high precision. All models except ResNet-50 maintained accuracy above 98\%, with confusion matrices revealing minimal misclassification patterns.

\begin{table}[ht]
    \centering
    \caption{Performance Metrics Across Embedding Models}
    \begin{tabular}{l c c c c}
        \hline
        \textbf{Models} & \textbf{Embedding} & \textbf{Precision} & \textbf{Recall} & \textbf{Accuracy} \\
        & \textbf{Dimension} & & & \\
        \hline
        CLIP & 512 & 0.9935 & 0.998 & 0.9951 \\
        Google VIT & 768 & 0.988 & 0.987 & 0.9856 \\
        DinoV2 & 768 & 0.9881 & 0.9945 & 0.9899 \\
        ResNet 50 & 2048 & 0.9898 & 0.826 & 0.8948 \\
        AIMv2 & 1024 & 0.992 & 0.994 & 0.9919 \\
        \hline
    \end{tabular}
    \label{tab:performance}
\end{table}

\subsection{Benchmarking different embedding models}
The goal of this experiment is to evaluate the robustness and consistency of different image embedding models when faced with manipulations in input images. By systematically modifying the original AI-generated images and comparing their embeddings with the unaltered AI images, the experiment answers if the embeddings differ entirely from their original embeddings, thus validating the embedding technique for image detection. We analyzed performance degradation across two key dimensions: geometric modifications and blur transformations. For geometric modifications, we applied both single and multiple white patch overlays and resolution reduction. 

The previous 9,000 AI-generated images were taken as original images; for every modification (6 total modifications), a set of 9000 modified images is stored separately. The vector embeddings of all the modified images using the five embedding models mentioned in experiment 1 are stored in different Pinecone indexes. The nearest distance of modified images from the original images (generated by AI) is calculated using cosine similarity and stored in spreadsheets. The closest distance is retrieved along with the metadata from the database using the vector similarity search that includes the name of the closest match image to compare whether the match returned the same image from the original dataset. The accuracy of the results for this evaluation is calculated as below.

\begin{equation}
\text{Accuracy} = \frac{\text{Number of Correct Matches}}{\text{Total Number of Images}} \times 100\%
\end{equation}

\text{where a correct match occurs when the modified image's}\\
\text{embedding is closest to its original AI image embedding}

The results as summarized in Table \ref{tab:geometric-performance} demonstrates that DINOv2 maintains exceptional resilience to geometric changes, achieving 100\% accuracy with single patches of 128 by 128 pixels and 99.94\% accuracy under resolution reduction to 128 by 128 pixels from 512 by 512 pixels. In contrast, multiple patch overlays presented a more significant challenge, particularly for the CLIP model, where accuracy decreased to 73.34\%.

\begin{table}[ht]
    \centering
    \caption{Model Accuracy (\%) for Patch Overlays and Resize}
    \begin{tabular}{l c c c c}
        \hline
        \textbf{Models} & \multicolumn{1}{c}{\textbf{Embedding}} & \textbf{1 Patch} & \textbf{3-5 Patches} & \textbf{Resize} \\
        & \textbf{Dimension} & & & \\
        \hline
        CLIP       & 512  & 97.26 & 73.34 & 98.04 \\
        Google VIT & 768  & 99.88 & 91.62 & 98.98 \\
        DinoV2     & 768  & 100.00 & 99.62 & 99.94 \\
        ResNet-50  & 2048 & 99.89 & 84.02 & 98.47 \\
        AIMV2      & 1024 & 99.96 & 92.59 & 99.81 \\
        \hline
    \end{tabular}
    \label{tab:geometric-performance}
\end{table}

The blur analysis reveals a systematic relationship between blur intensity and model performance, as detailed in Table \ref{tab:blur_performance}. At low blur levels (20\%), most models maintained robust performance, with DinoV2 achieving perfect accuracy and both Google ViT and AIMv2 exceeding 99\% accuracy. However, as the blur intensity increased, we can observe a consistent degradation pattern across all models, though at varying rates. DinoV2 demonstrated remarkable resilience, maintaining 88.44\% accuracy even at 80\% blur intensity, significantly outperforming other models. In contrast, ResNet-50, despite its larger embedding dimension of 2048, shows the highest sensitivity to blur effects, with accuracy dropping dramatically to 24.67\% at 80\% blur.

\begin{table}[ht]
    \centering
    \caption{Model Accuracy (\%) Under Various Blur Intensities}
    \begin{tabular}{l c c c c c}
        \hline
        \textbf{Models} & \textbf{Embedding} & \multicolumn{4}{c}{\textbf{Blur Intensity}} \\
        \cline{3-6}
        & \textbf{Dimension} & \textbf{20\%} & \textbf{40\%} & \textbf{60\%} & \textbf{80\%} \\
        \hline
        CLIP & 512 & 96.68 & 80.96 & 48.12 & 30.02 \\
        Google ViT & 768 & 99.86 & 96.00 & 84.17 & 67.06 \\
        DinoV2 & 768 & 100.00 & 99.87 & 96.86 & 88.44 \\
        ResNet-50 & 2048 & 98.52 & 74.18 & 44.36 & 24.67 \\
        AIMv2 & 1024 & 99.65 & 96.81 & 90.08 & 80.13 \\
        \hline
    \end{tabular}

    \label{tab:blur_performance}
\end{table}

Our comprehensive evaluation yielded several insights for the field. The exceptional performance across multiple embedding models suggests that AI-generated images possess consistent, detectable patterns in the embedding space, with transformer-based models generally outperforming traditional CNN architectures like ResNet 50. Also, larger embedding dimensions did not necessarily correlate with better performance, while model architecture significantly influenced robustness against different types of modifications.

\subsection{Smart Contract Gas Usage Estimation}

This experiment was designed to evaluate and compare the gas consumption incurred when storing 256-bit embedding hashes on the Ethereum blockchain using two different Solidity data types: \texttt{uint256} and \texttt{string}. The purpose was to empirically determine the gas efficiency of each approach, which is critical when scaling decentralized applications that store or verify large volumes of hashed data.

To empirically assess gas usage differences, two separate Solidity smart contracts were implemented:
\begin{itemize}
    \item \textbf{HashStorageInt}: Stores the 256-bit image embedding hash as \texttt{uint256}.
    \item \textbf{HashStorageStr}: Stores the same 256-bit hash as a \texttt{string}.
\end{itemize}

Each smart contract included two core functions: \texttt{storeHash(input\_hash)} for storing a hash value on the blockchain, and \texttt{hashExists(input\_hash)} for verifying whether a given hash is already present in the contract’s storage.

Both contracts were developed and tested in a local Hardhat Ethereum environment, where extensive unit tests were written to validate the functionality. Subsequently, both contracts were deployed on the Sepolia testnet, where transactions were executed to measure real-world gas usage. Only 200 image embedding hashes were stored on Sepolia due to constraints in transaction time and limited gas availability.

Gas usage was recorded for each transaction involving the \texttt{storeHash (input\_hash)} function in both contracts. The summarized statistics are provided in Table~\ref{tab:gas_usage}. These findings clearly show that storing a 256-bit embedding hash as a \texttt{uint256} is approximately 2.7 times more gas efficient than storing it as a \texttt{string}.

\begin{table}[ht]
    \centering
    \caption{Gas Usage Statistics for Hash Storage Using \texttt{uint256} and \texttt{string}}
    \begin{tabular}{l c c}
        \hline
        \textbf{Statistic} & \textbf{Gas (uint256)} & \textbf{Gas (string)} \\
        \hline
        Count   & 9,000 transactions & 9,000 transactions \\
        Mean    & 36,207 gas         & 97,667 gas \\
        Median  & 36,184 gas         & 97,667 gas \\
        Minimum & 21,528 gas         & 97,667 gas \\
        Maximum & 51,228 gas         & 97,667 gas \\
        \hline
    \end{tabular}
    \label{tab:gas_usage}
\end{table}

\subsection{Framework Evaluation}
To assess the practical performance of our AI image detection system, we developed and deployed a working prototype capable of performing image classification based on stored image embeddings and integration of smart contracts. The goal of this evaluation is to understand how the system behaves when given input images from the real world and whether it can accurately categorize images as AI-generated or human-generated. The prototype allows for uploading any image, processes it through an embedding model, and returns a classification result based on the similarity match with the embeddings stored in the local database.

This evaluation phase focused not only on the overall classification accuracy, but also on identifying potential limitations, especially in diverse and uncontrolled image settings. The prototype was subjected to various image categories to simulate practical use cases. We specifically examined how the system handles facial images (often generated by StyleGAN) compared to non-facial or generic image categories, such as artwork and nature.

\subsubsection{Framework 1: Blockchain only Evaluation}
To evaluate the blockchain-only prototype, we conducted manual testing using the system's user interface by uploading individual image files for classification. In this setup, the backend system used a predefined dataset of 7,000 AI-generated images and 7,000 real (human) images. For each image, an embedding was generated using the DINOv2 model, and a 256-bit cryptographic hash of this embedding was computed and stored on the Ethereum Sepolia test network via two dedicated smart contracts: \texttt{HashStorageAI} for AI-generated images and \texttt{HashStorageHuman} for real ones. When a new image is uploaded, the system computes its embedding and hash, then compares the hash against both smart contracts. If a match is found in either contract, the image is classified accordingly—AI or human—and marked as verifiable. Otherwise, if no match exists, the system cannot make a definitive classification, and the result is labeled as inconclusive.

Although functionally sound in design, this evaluation method encountered several practical challenges due to limitations inherent in on-chain data storage. Specifically, storing all 14,000 image hashes on the Ethereum blockchain introduced high computational and financial overhead. Every transaction to add a hash to a smart contract incurs a gas fee, and when many transactions are sent in rapid succession, network congestion and rate limitations often lead to failed or dropped transactions. From a technical standpoint, this results in nonce conflicts, out-of-gas errors, or rate-limiting by the node provider (Alchemy), thereby preventing the successful storage of all intended hashes. Consequently, while every image embedding was stored in the local vector database, only a subset of those corresponding hashes could be reliably registered on-chain. As a result, many uploaded images during evaluation matched an embedding in the database but failed the blockchain verification check—leading to false negatives in verifiability. This limitation highlights the trade-offs between blockchain transparency and system scalability, particularly in use cases involving large-scale data operations.

\subsubsection{Framework 2: Database only Evaluation}
\begin{figure}
    \centering
    \includegraphics[width=0.7\linewidth]{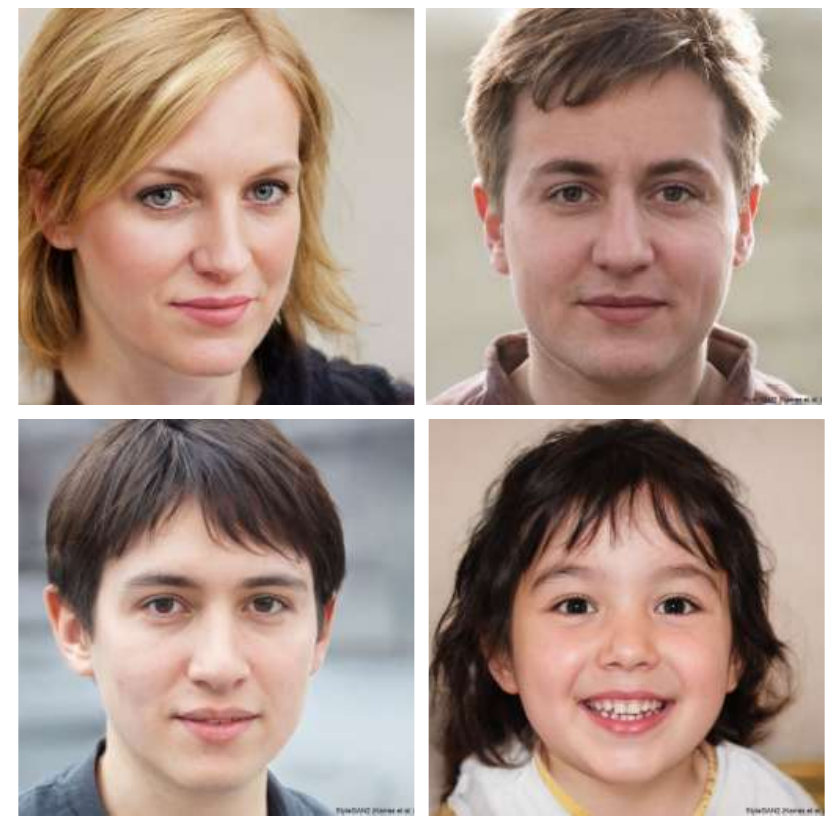}
    \caption{StyleGAN-generated synthetic images used for EmbedAIDetect prototype evaluation}
    \label{fig:style-gan}
\end{figure}

The second prototype focused solely on embedding-based classification using vector similarity search. Before formally evaluating Prototype 2, we expanded the system's internal database to better support classification of facial images—an area where the system initially struggled. We first introduced 10,000 high-resolution AI-generated facial images created using StyleGAN, sourced from \url{https://thispersondoesnotexist.com/}, into the AI category. The model showed strong performance in identifying similar StyleGAN images during testing, correctly labeling over 90\% of such inputs as synthetic. Some
examples of correctly classified AI faces are shown in Figure~\ref{fig:style-gan}. However, this also led to a growing tendency to misclassify real human faces as AI-generated, suggesting the embedding space was becoming biased toward synthetic features.

To mitigate this issue, we integrated 1,441 real facial images from the Chicago Face Database (CFD)~\cite{ma2015cfd, ma2020cfdmr, lakshmi2020} into the human category. This addition provided more diversity in real facial representations and slightly improved accuracy for some real images during early tests. However, the underlying issue persisted: the model continued to misclassify many real human faces, particularly those with uniform lighting and frontal poses, likely because of their visual similarity to StyleGAN outputs. This imbalance in the database influenced how the system later performed during the formal evaluation stage.

With the internal database prepared, we proceeded to evaluate the prototype using two external datasets: (1) CIFAKE~\cite{bird2024cifake, cifake2023}, which includes a wide range of non-facial real and AI-generated images, and (2) 140k Real and Fake Faces~\cite{realfakefaces2021}, which contains high-resolution facial images. The goal was to evaluate how accurately the system could classify input images as AI-generated or real based on similarity to pre-stored embeddings using cosine distance. The system was tested using Python scripts that emulated the classification flow of the application, including embedding computation, vector similarity querying, and prediction logging. Results were saved in CSV format with fields such as filename, true label, similarity scores, and predicted label. A snippet of this output is shown in Table~\ref{tab:sample_predictions}.

\begin{table}[ht]
\scriptsize
\centering
\caption{Sample Prediction Output}
\begin{tabular}{l c c c c}
\hline
\textbf{Filename} & \textbf{True Label} & \textbf{Human Sim.} & \textbf{AI Sim.} & \textbf{Pred. Label} \\
\hline
0375 (7).jpg & real & 0.2266 & 0.1677 & real \\
0838 (7).jpg & real & 0.2846 & 0.2476 & real \\
0174 (5).jpg & real & 0.4529 & 0.5668 & fake \\
0445 (3).jpg & real & 0.1606 & 0.1622 & fake \\
0771 (9).jpg & real & 0.2502 & 0.3689 & fake \\
\hline
\end{tabular}
\label{tab:sample_predictions}
\end{table}

We then computed standard classification metrics such as accuracy, precision, recall, and F1-score, along with confusion matrices, to assess how the system performed across both facial and non-facial image categories.

\subsubsection{Quantitative Results and Observations}
\begin{figure}[ht]
    \centering
    
    \begin{subfigure}[t]{0.48\columnwidth}
        \includegraphics[width=\linewidth]{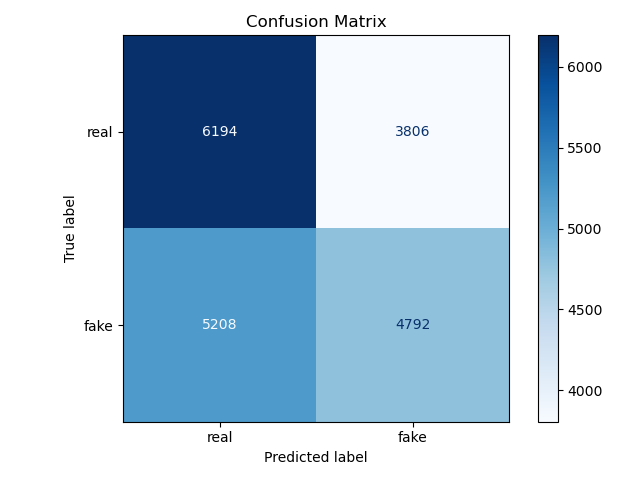}
        \caption{CIFAKE dataset}
        \label{fig:cifake-matrix}
    \end{subfigure}
    \hfill
    \begin{subfigure}[t]{0.48\columnwidth}
        \includegraphics[width=\linewidth]{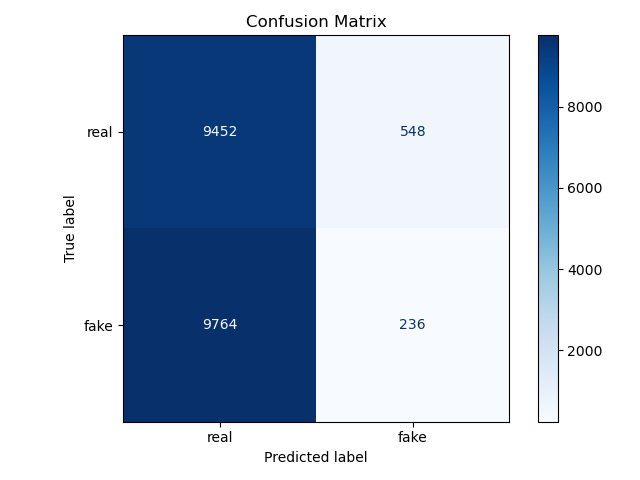}
        \caption{140k Faces dataset}
        \label{fig:faces-matrix}
    \end{subfigure}

    \caption{Confusion matrices showing classification performance of Prototype 2 using vector similarity search}
    \label{fig:prototype-eval-2}
\end{figure}

On the CIFAKE dataset, which contains objects, scenes, and non-facial artwork, the system showed balanced performance. As shown in Figure~\ref{fig:cifake-matrix}, it correctly classified 6,194 out of 10,000 real images and 4,792 out of 10,000 fake images. While not highly accurate, the results indicate that the model can reasonably distinguish between AI-generated and real content in general image categories where visual patterns are more distinct and less uniform.

In contrast, the evaluation on the 140k Real and Fake Faces dataset highlighted significant limitations. As shown in Figure~\ref{fig:faces-matrix}, the system correctly labeled most real facial images (9,452 out of 10,000), but misclassified nearly all fake faces: only 236 were correctly identified as AI-generated, while 9,764 were incorrectly predicted as real. This imbalance reflects a strong bias toward labeling facial content as real, likely due to the embedding model’s inability to capture subtle generative artifacts present in StyleGAN outputs, especially when those features overlap with realistic cues found in the CFD dataset.

These results reinforce a key insight: Although embedding-based similarity is reasonably effective for broad semantic differences (as seen in non-facial images), it falls short when used to detect highly realistic facial forgeries. This gap directly influenced the decision to incorporate blockchain-based verification in the hybrid prototype, with the aim of enhancing confidence and trust in high-risk classifications.

\subsubsection{Framework 3: Hybrid Approach Evaluation}
Prototype 3 combines the embedding-based similarity classification from Prototype 2 with the blockchain verification mechanism from Prototype 1. As it does not introduce new classification logic, but rather integrates existing components sequentially, first identifying the closest match via vector similarity, then checking its hash on-chain, no separate evaluation was conducted. Its behavior and performance characteristics are directly inherited from the earlier prototypes. The primary goal of this version is to enhance trust and verifiability in the results, especially in edge cases, by confirming that the matched embedding has an immutable record on the blockchain. Although not formally benchmarked, the hybrid workflow was manually validated through the user interface and confirmed to work as expected.

The results suggest that the system accuracy increases with the diverse data set, and important consideration includes necessary trade-offs between processing efficiency and accuracy. Our paper presents a concept that can be further used with other image analysis techniques to determine the authenticity of an image.  This research has found that DINOv2's superior robustness makes it particularly suitable for applications where image modifications are expected. Several areas relating to image identification warrant further investigation, including the impact of different AI generation models on detection accuracy, effectiveness against adversarial modifications, scalability with larger datasets, and performance against more sophisticated image modifications.

\section{Conclusion}\label{sec:conclusion}
The research paper presents a novel concept of identifying and detecting AI-generated images. The findings suggest that despite extensive manipulations, the semantic meaning of images remains similar to their original counterparts, allowing for identification. However, the efficacy of this approach is inherently limited by the dataset used for embedding generation. Also, the widespread adoption of this system across big technology companies such as OpenAI, Meta, and Google is crucial to ensuring its success. Standardization and a shared database of vector embeddings are necessary because different embedding models produce varying representations. Future work will focus on leveraging blockchain and IPFS to establish a decentralized image verification system. By ensuring the integrity and accessibility of vector embeddings, such a system could pave the way for a more robust and universally accepted AI image detection framework. Ultimately, collaboration across the industry will be vital in creating a scalable, transparent, and effective solution for combating the challenges posed by AI-generated media.
% \printbibliography

\bibliographystyle{ieeetr}
\bibliography{references}

\end{document}